\documentclass[acmlarge]{acmart}
\AtBeginDocument{%
  }

\usepackage{multirow}
\usepackage{graphicx} 
\usepackage{subcaption} 
\usepackage{xcolor} 
\usepackage{caption}

\setcopyright{none}
\makeatletter
\renewcommand{\@copyrightowner}{}
\renewcommand{\@copyrightpermission}{}
\renewcommand{\@formatdoi}[1]{}
\renewcommand{\@copyrightyear}{}
\makeatother


\makeatletter
\def\@acmReferenceFormat#1{} 
\def\@acmPages{} 
\makeatother

\begin{document}

\title{Explainable, Multi-modal Wound Infection Classification from Images Augmented with Generated Captions}

\author{Palawat Busaranuvong}

\email{pbusaranuvong@wpi.edu}
\author{Emmanuel Agu}
\authornote{Corresponding author.}
\email{emmanuel@wpi.edu}
\author{Reza Saadati Fard}
\email{rsaadatifard@wpi.edu}
\author{Deepak Kumar}
\email{dkumar1@wpi.edu}
\author{Shefalika Gautam}
\email{sgautam@wpi.edu}
\author{Bengisu Tulu}
\email{bengisu@wpi.edu}
\author{Diane Strong}
\email{dstrong@wpi.edu}

\affiliation{%
  \institution{Worcester Polytechnic Institute}
  \city{Worcester}
  \state{MA}
  \country{USA}
}

\renewcommand{\shortauthors}{Busaranuvong et al.}

\begin{abstract}
Infections in Diabetic Foot Ulcers (DFUs) can cause severe complications, including tissue death and limb amputation, highlighting the need for accurate, timely diagnosis. Previous machine learning methods have focused on identifying infections by analyzing wound images alone, without utilizing additional metadata such as medical notes. In this study, we aim to improve infection detection by introducing \underline{S}ynthetic \underline{C}aption  \underline{A}ugmented \underline{R}etrieval for \underline{W}ound \underline{I}nfection \underline{D}etection (SCARWID), a novel deep learning framework that leverages synthetic textual descriptions to augment DFU images. SCARWID consists of two components: (1) Wound-BLIP, a Vision-Language Model (VLM) fine-tuned on GPT-4o-generated descriptions to synthesize consistent captions from images; and (2) an Image-Text Fusion module that uses cross-attention to extract cross-modal embeddings from an image and its corresponding Wound-BLIP caption. Infection status is determined by retrieving the top-$k$ similar items from a labeled support set. To enhance the diversity of training data, we utilized a latent diffusion model to generate additional wound images. As a result, SCARWID outperformed state-of-the-art models, achieving average sensitivity, specificity, and accuracy of 0.85, 0.78, and 0.81, respectively, for wound infection classification. Displaying the generated captions alongside the wound images and infection detection results enhances interpretability and trust, enabling nurses to align SCARWID outputs with their medical knowledge. This is particularly valuable when wound notes are unavailable or when assisting novice nurses who may find it difficult to identify visual attributes of wound infection.
\end{abstract}



\settopmatter{printacmref=false}
\maketitle

\section{Introduction}
\label{sec:introduction}

Chronic wounds present a considerable health challenge in the United States, impacting over 6.5 million individuals (2\% of the population)~\cite{jarbrinkhumanistic}. Affecting predominantly older adults~\cite{nussbaum2018economic, gould2015chronic}, these wounds severely impact the patients' quality of life and impose a significant financial burden, with annual medicare expenditures ranging from \$28.1 to \$96.8 billion~\cite{sen2009human}. 
Complications often arise due to infections, which can require emergency interventions and potentially lead to limb amputation if not addressed promptly~\cite{partb_DFU}. This paper addresses infection classification in Diabetic Foot Ulcers (DFUs), which are especially dangerous for individuals with diabetes. More than half of all DFUs become infected, leading to amputations in 20\% of cases at a cost of \$33,499 per amputation \cite{olsson2019humanistic, mills2014society}.

In current medical practice, diagnosing an infected wound involves several steps: debridement (removal of dead tissues), blood tests, and expert evaluation, which are typically conducted in a clinical setting \cite{lipsky2016antimicrobial, maclellan2002designing, stallard2018and}. This protocol presents challenges at the Point of Care (POC), such as in patients' homes or at trauma sites, where before debridement, non-specialist caregivers may suspect an infection but do not have access to specialty services to follow the protocol. Often, these caregivers must advise patients to seek further evaluation at a clinic or emergency facility to confirm the presence of an infection. This referral process not only delays treatment but also increases the risk of severe outcomes, including amputations \cite{rondas2015prevalence}. Furthermore, many wounds that are referred for expert assessment are subsequently found to be uninfected, leading to unnecessary use of resources such as transportation and additional costs such as emergency department charges \cite{wilbright2004use, chanussot2013telemedicine}. In settings where clinicians do not have access to detailed clinical data, they are forced to rely on visual inspections to spot early signs of infection in Diabetic Foot Ulcers (DFUs), such as increased redness, swelling, warmth, and the presence of colored purulent discharge. However, these inspections are not always accurate and are challenging for nurses or caregivers with insufficient wound experience.

\begin{figure*}[!th]
\centering
  \includegraphics[width=0.95\textwidth]{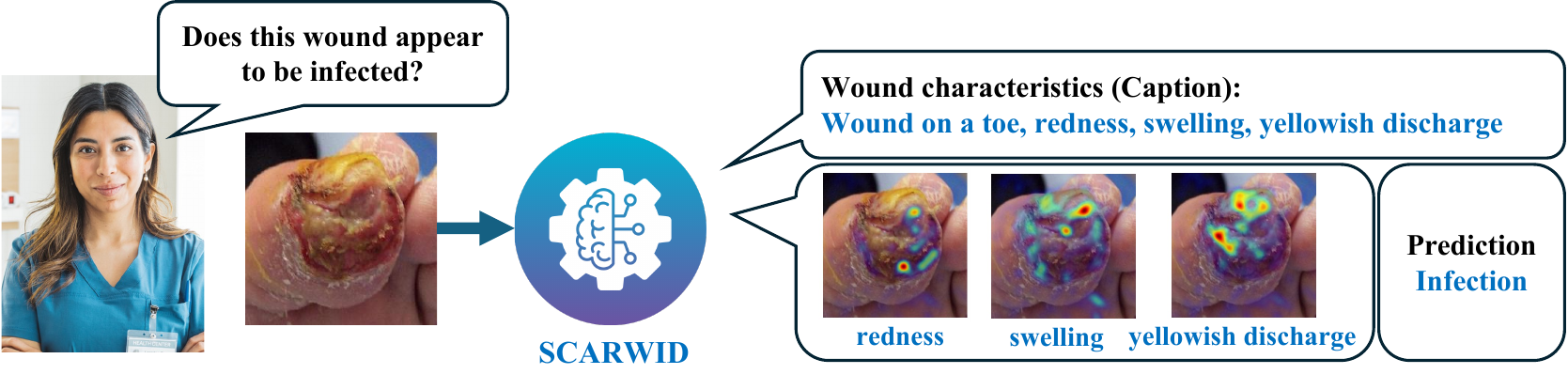}
  \caption{A comprehensive solution for identifying infection in wound images along with annotations}
  \label{fig:solution}
\end{figure*}

Despite the application of State-of-the-art (SOTA) deep learning models ~\cite{wang2015unified, partb_DFU, al2022diabetic, ConDiff, yap2021analysis, qayyum2021vit, galdran2021convolutional} in classifying infections from visual appearances of wounds in photographs without relying on wound tests, medical notes, or extensive clinical examinations, 
they still lack interpretability for humans, particularly in explaining why a wound is flagged as infected or uninfected. Novice nurses often find it challenging to identify attributes of a wound that suggest that it is infected, even with abundant wound care decision guidelines~\cite{franks2016management}. Therefore, to build trust in wound assessment systems among nurses, deep learning frameworks designed for novice wound care providers must include clear explanations and annotations that highlight which visual characteristics of a wound in an image indicate the presence or absence of infection.

\textbf{Our approach}: To address these issues, we propose a comprehensive deep learning model (see Fig.~\ref{fig:solution}) that improves the accuracy of wound infection prediction over SOTA models with enhanced interpretability. This paper introduces an integrated framework that combines a Vision-Language pre-trained model with a multimodal classification model, termed \underline{S}ynthetic \underline{C}aption  \underline{A}ugmented \underline{R}etrieval for \underline{W}ound \underline{I}nfection \underline{D}etection or SCARWID, for the classification of infections in DFU photographs.
Our approach involves generating visual highlights and annotations of the wound image along with textual descriptions of wound characteristics from an input image to help novice nurses understand the wound's attributes indicative of infection. By reflecting on both the wound image and the corresponding textual description of infection attributes, the SCARWID model's rationale for infection classification can more easily be understood, potentially improving a nurse's wound expertise in the longer term. 

Due to the absence of medical notes corresponding to the DFU images in our dataset, we employed GPT-4o~\cite{gpt4}, a Multimodal Large Language Model (MLLM) capable of processing both text and visual data making it highly effective in tasks such as image captioning, to generate concise descriptions of wound images. These captions, highlighting potential signs of infection, served as metadata for training the SCARWID model. We provided expert-labeled wound statuses (infected or uninfected) to GPT-4o to guide the caption generation process and refine the descriptions reflecting the wound's condition. Next, we fine-tuned a Vision-Language Pre-training model known as \textit{Bootstrapping Language-Image Pre-training} (BLIP)~\cite{blip} on the image captioning task with GPT-4o generated descriptions. This resulted in a captioning model called Wound-BLIP that provides consistent descriptions without needing label information at test time. This method not only enriched our dataset but also deepened our understanding of the rationale behind expert labeling decisions for wound images.

During inference, our proposed SCARWID model classifies infections by processing a DFU image query along with its corresponding wound description generated by Wound-BLIP. The model retrieves the top $k$ most similar image-text pairs from a database of labeled support documents, where similarity is determined based on the closest distance in the embedding space. By default, k=5. The final prediction is made by selecting the most frequently occurring label among these $k$ retrieved items.

\textbf{Our main contributions} are as follows:

\begin{itemize}
   \item We propose SCARWID, an integrated end-to-end framework, which combines wound images with their descriptions to transform them into multimodal embeddings. Classifications are made based on the most common labels among the top $k$ most similar pairs of images and texts retrieved from the support database.

  \item We fine-tuned the BLIP image-captioning model using 1,000 pairs of DFU images and their corresponding text generated by GPT-4o
   , facilitating the generation of textual meta-data essential for infection classification.

   \item SCARWID was evaluated on 5-fold cross-validation protocol and demonstrated significant improvements of 4-9\% in sensitivity and specificity over SOTA wound image classification models such as CNN-Ensemble and DFU-RGB-TEX-Net. Furthermore, it demonstrated high robustness and generalization evidenced by the lower standard deviations of evaluation scores. 
   
    \item To enhance interpretability, we present examples of SCARWID's predictions with visual highlights, annotations, and corresponding textual wound descriptions. Specifically, for sample wounds, we concurrently display: (1) wound regions highlighted by Grad-CAM on the Wound-BLIP image-ground text encoder, showing where descriptions of specific wound characteristics are most evident; and (2) image attributes that our image-text fusion module focuses on when retrieving similar images from the support database, visualized using attention heatmaps.
\end{itemize}

\section{Related Work}
\label{sec:RelatedWork}

\subsection{Wound Infection Classification with Deep Learning}

\begin{table}[ht]
\centering
\caption{Summary of prior work on wound infection classification using deep learning}
\resizebox{\textwidth}{!}{
\begin{tabular}{cccccc}
\hline
\textbf{Specific ML problem} &
  \textbf{Related Work} &
  \textbf{\begin{tabular}[c]{@{}c@{}}Summary of \\ Approach\end{tabular}} &
  \textbf{\begin{tabular}[c]{@{}c@{}}No. of \\ Target Classes\end{tabular}} &
  \textbf{Dataset} &
  \textbf{Results} \\ \hline
\centering
\begin{tabular}[c]{@{}c@{}}Wound segmentation \\ and Infection \\ Classification\end{tabular} &
  \begin{tabular}[c]{@{}c@{}}Wang et al. \\ 2015 \cite{wang2015unified}\end{tabular} &
  \begin{tabular}[c]{@{}c@{}}CNN-based: \\ ConvNet + SVM\end{tabular} &
  \begin{tabular}[c]{@{}c@{}}2 classes \\ (infection and\\  no infection)\end{tabular} &
  \begin{tabular}[c]{@{}c@{}}NYU wound \\ Database\end{tabular} &
  \begin{tabular}[c]{@{}c@{}}Accuracy: 95.6\%\\ PPV: 40\%\\ Sensitivity: 31\%\end{tabular} \\ \hline 
\centering
 &
  \begin{tabular}[c]{@{}c@{}}Goyal et al.\\ 2020 \cite{partb_DFU} \end{tabular} &
  \begin{tabular}[c]{@{}c@{}}CNN-based: \\ Ensemble CNN\end{tabular} &
  \multirow{3}{*}{\begin{tabular}[c]{@{}c@{}} \\ \\ 2 classes \\ (infection and \\ no infection)\end{tabular}} &
  \multirow{3}{*}{\textit{\begin{tabular}[c]{@{}c@{}} \\ Part B DFU \\  2020 dataset\\ \\ (We also used \\ this dataset)\end{tabular}}} &
  \begin{tabular}[c]{@{}c@{}}Accuracy: 72.7\%\\ PPV: 73.5\%\\ Sensitivity: 70.9\%\end{tabular} \\ \cline{2-3} \cline{6-6} 
\centering
\multirow{-2}{*}{\begin{tabular}[c]{@{}c@{}}DFU infection \\ classification\end{tabular}} &
  \begin{tabular}[c]{@{}c@{}}Al-Garaawi et al. \\ 2022 \cite{al2022diabetic} \end{tabular} &
  \begin{tabular}[c]{@{}c@{}}CNN-based:\\ DFU-RGB-TEX-Net\end{tabular} &
   &
   &
  \begin{tabular}[c]{@{}c@{}}Accuracy: 74.2\%\\ PPV: 74.1\%\\ Sensitivity: 75.1\%\end{tabular} \\ \cline{2-3} \cline{6-6} 
\centering
 &
   \begin{tabular}[c]{@{}c@{}}Busaranuvong et al. \\ 2024 \cite{ConDiff} \end{tabular} &
  \begin{tabular}[c]{@{}c@{}}Generative-Discrimination:\\ ConDiff (Distance-based)\end{tabular} &
   &
   &
  \begin{tabular}[c]{@{}c@{}} Accuracy: 83.3\%\\ PPV: 85.8\%\\ Sensitivity: 85.8\%\end{tabular} \\ \hline
\centering
\multirow{3}{*}{\begin{tabular}[c]{@{}c@{}}DFU wound ischemia\\ and  infection \\ classification\end{tabular}} &
  \begin{tabular}[c]{@{}c@{}}Yap et al. \\ 2021 \cite{yap2021analysis} \end{tabular} &
  \begin{tabular}[c]{@{}c@{}}CNN-based:\\ VGG, ResNet,\\ InceptionV3, DenseNet, \\ EfficientNet\end{tabular} &
  \multirow{3}{*}{\begin{tabular}[c]{@{}c@{}}4 classes\\ (both infection\\ and ischemia,\\ infection, ischemia,\\ none)\end{tabular}} &
  \multirow{3}{*}{\begin{tabular}[c]{@{}c@{}} \\  DFUC2021\\ dataset\end{tabular}} &
  \begin{tabular}[c]{@{}c@{}}EfficientNet B0 \\ performance: \\ F1, PPV, SEN\\ =  55\% , 57\%, 62\%\end{tabular} \\ \cline{2-3} \cline{6-6} 
\centering
 &
  \begin{tabular}[c]{@{}c@{}}Qayyum et al. \\ 2021 \cite{qayyum2021vit} \end{tabular} &
  \begin{tabular}[c]{@{}c@{}}ViT-based: \\
  Ensemble ViT \end{tabular} &
   &
   &
  \begin{tabular}[c]{@{}c@{}} F1, PPV, SEN\\ = 57\%, 58\% , 61\%\end{tabular} \\ \cline{2-3} \cline{6-6} 
\centering
 &
  \begin{tabular}[c]{@{}c@{}}Galdran et al. \\ 2021 \cite{galdran2021convolutional} \end{tabular} &
  \begin{tabular}[c]{@{}c@{}}ViT-based: ViT, DeiT 
  \\ CNN-based: BiT, \\ EfficientNet\end{tabular} &
   &
   &
  \begin{tabular}[c]{@{}c@{}}BiT performance:\\ F1, PPV, SEN\\ = 61\%, 61\% , 66\%\end{tabular} \\ \hline
\end{tabular}
}
\label{tab:summary_of_approaches}
\end{table}

State-of-the-art (SOTA) Deep learning models that detect infections from wound images have become increasingly prevalent~\cite{partb_DFU, galdran2021convolutional, yap2021analysis, al2022diabetic}. Goyal et al.~\cite{partb_DFU} introduced the Part-B DFU dataset, which includes wound images for a infection classification task from diabetic foot ulcers. 
As detailed in Table~\ref{tab:summary_of_approaches}, Goyal et al.~\cite{partb_DFU} employed a CNN ensemble model that combines bottleneck features from CNN architectures and classifies using an SVM classifier, achieving 70.9\% sensitivity and 74.4\% specificity in binary infection classification. In subsequent research, Al-Garaawi et al.~\cite{al2022diabetic} developed a custom CNN framework, DFU-RGB-TEX-Net, which enhances feature extraction from DFU images using mapped binary patterns. DFU-RGB-TEX-Net integrates a linear combination of the original image and texture information as input for a CNN, resulting in a sensitivity of 75.1\% and a specificity of 73.4\%.

Busaranuvong et al.~\cite{ConDiff} proposed the ConDiff model for the classification of wound infections. ConDiff uses distance-based classification to predict the wound status based on the similarity between an input image and image-guided conditional synthetic images generated from infection and non-infection labels. ConDiff outperformed other SOTA models achieving 85.4\% sensitivity and 74.7\% specificity on the Part-B DFU infection dataset, demonstrating the potential of distance-based classification of wound imaging tasks. However, the downside of the ConDiff approach is its high computational cost during inference (4-5 seconds per image on an NVIDIA A100 GPU) due to the image-generating time with the diffusion model. 
This work also showed that more recent Vision Transformer (ViT)-based models such as SwinV2~\cite{swinv2} (82.7\% sensitivity and 69.8\% specificity) and EfficientFormer~\cite{efficientformer} (84.1\% sensitivity and 69.2\% specificity) outperformed CNN-based models in wound infection classification.

Galdran et al.~\cite{galdran2021convolutional} and Qayyum et al.~\cite{qayyum2021vit} explored SOTA ViT-based models for multiclass classification of ischemia and infection using the DFUC2021 challenge dataset provided by Yap et al.~\cite{yap2021analysis}. Their findings demonstrated that the performance of ViT-based was comparable to that of traditional CNN-based models on this task. Specifically, a ViT ensemble model~\cite{qayyum2021vit} achieved a sensitivity of 61\% and a positive predictive value (PPV) of 58\%, while the Big Transfer (BiT) model~\cite{galdran2021convolutional} achieved a sensitivity of 66\% and a PPV of 61\%.

\subsection{Medical Visual Question Answering with Multimodal Large Language Models}

LLMs have been explored for their proficiency in medical tasks. Models such as Med-PaLM~\cite{medPALM}, Med-PaLM2~\cite{medPALM2}, and GPT-4~\cite{GPT4_91acc} achieve impressive accuracies of 67.6\%, 86.5\%, and 90.1\%,  respectively on multiple-choice US Medical Licensing Examination (USMLE) questions, well above the exam's approximate passing score of 60\% \cite{usmle_score}.

Despite these advancements, challenges persist for the Medical Visual Question Answering (medical VQA) task. For example, while Med-PaLM2 excels in text-based analysis, it lacks visual data interpretation capabilities. In contrast, GPT-4o, a Multimodal Large Language Model (MLLM), effectively integrates visual and textual information. Jin et. al~\cite{jin2024hidden} shows that GPT-4o achieves an accuracy of 88\% in the New England Journal of Medicine (NEJM) Image Challenge when medical images and clinical information are provided, outperforming the average physician's accuracy of 77\%. This finding is in line with another experiment~\cite{GPT4V_with_hint}, which illustrates that incorporating expert hints into the USMLE with image questions taken from the AMBOSS medical platform increases the accuracy of GPT-4o from 60-68\% to 84-88\%, highlighting its potential for improved medical diagnostic support.

However, GPT-4o's performance drops significantly in the NEJM image challenge scenarios where only medical images are used as inputs, with diagnostic accuracy ranging from 29-40\%, and accuracy around 42-50\% when only providing essential information about the patient, their symptoms, and relevant clinical details~\cite{GPT4V_29acc, GPT4V_61acc}.  
This highlights a critical gap in its ability to process purely visual information without supporting context from text or other modalities.

In our research, we focus on infection classifications from wound images since prioritizing infection detection is crucial for addressing urgent clinical requirements and enabling timely and appropriate treatment interventions, such as the initiation of antibiotic therapy or surgical procedures. Our paper addresses scenarios in which additional patient clinical information, medical notes, or descriptions corresponding to each DFU image are unavailable. As mentioned above, using GPT-4o to analyze only wound images for infection classification is not recommended. As an alternate strategy, we address GPT-4o's limitations in image-only analysis by incorporating expert labels of DFU images to generate wound descriptions. Later, these descriptions are used for fine-tuning the BLIP image captioning model that generates wound image descriptions without using unavailable expert-assigned labels at test time.

\section{Methodology}
\label{sec:methodology}


 \begin{figure}[!th]
  \centering
  \includegraphics[width=0.95\textwidth]{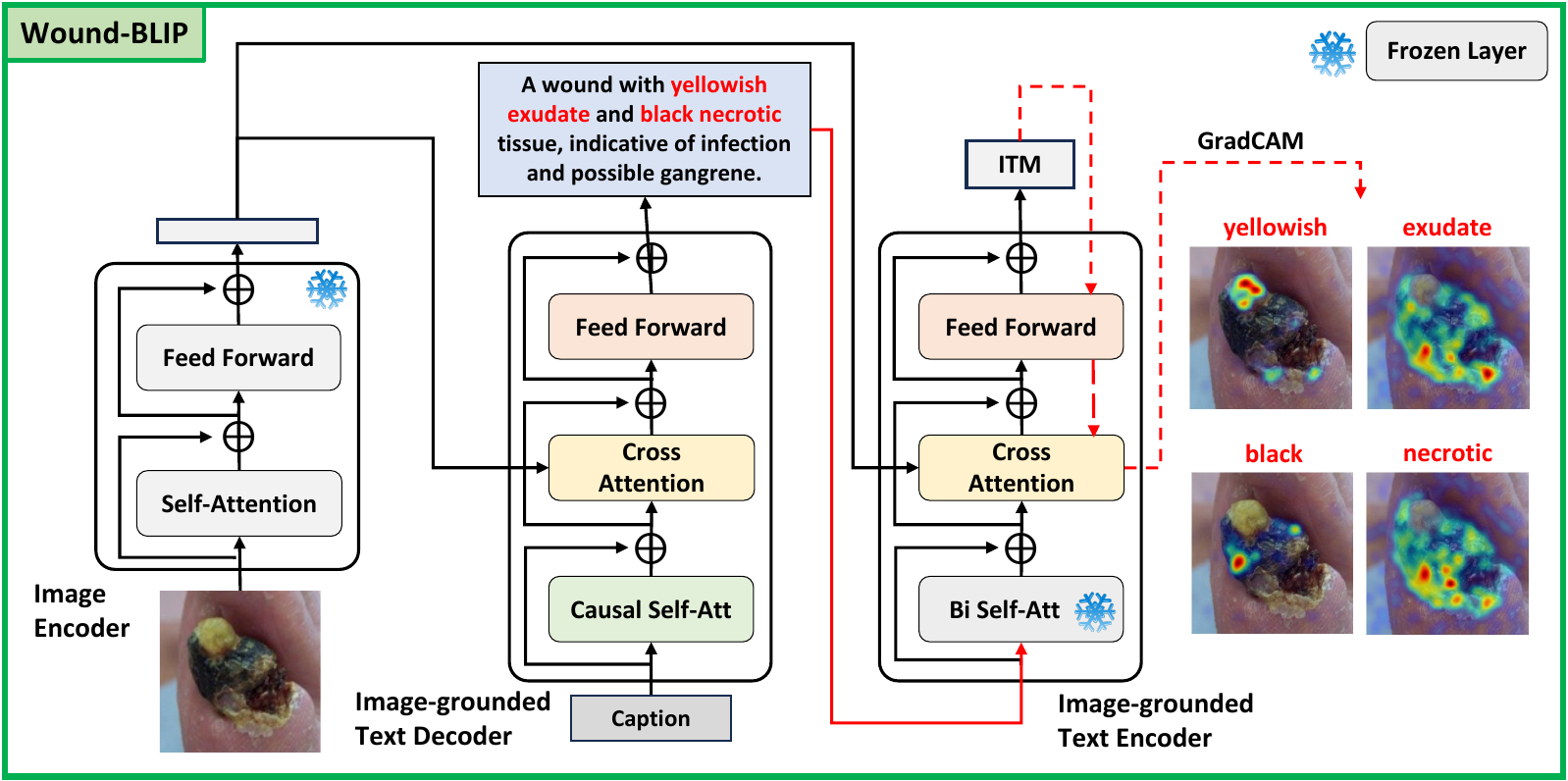}
  \caption{\textbf{Overview of the Wound-BLIP Architecture.} The model uses a wound Image Encoder to process a wound image and then uses an Image-grounded Text Decoder to generate a concise description of the wound. To enhance interpretability, an Image-grounded Text Encoder is utilized to visualize text localization via Grad-CAM heatmaps based on synthetic wound descriptions.}
  \label{fig:wound_blip}
\end{figure}

\subsection{Wound-BLIP Image Captioning Model} \label{sec:VLM}

Vision-Language Models (VLMs) are designed to understand and generate information from both visual and textual data. They can analyze images and relate them to corresponding text, enabling outputs such as captions, answers to questions about visual content, or textual summaries of scenes. Examples of VLMs include BLIP~\cite{blip}, BLIP-2~\cite{blip2}, Flamingo~\cite{flamingo}, and LLaVA~\cite{visual_instruct}.

We selected BLIP~\cite{blip} as the image captioning model to generate textual descriptions of wounds from images because it is smaller in size compared to other VLM approaches. Unlike models that use a large language model (LLM) backbone as a text decoder, BLIP uses an image-grounded text decoder that can be easily fine-tuned. Since our downstream task is to describe characteristics of wounds from images, we refer to our fine-tuned BLIP model as \textit{Wound-BLIP}.

The Wound-BLIP architecture for wound image captioning consists of three main components: (1) an Image Encoder, (2) an Image-grounded Text Decoder, and (3) an Image-grounded Text Encoder (see Fig.\ref{fig:wound_blip}). The Image Encoder processes the input image into a sequence of embeddings that capture the contextual relationships within the image, utilizing a Vision Transformer (ViT) architecture\cite{vit}.

\subsubsection{Image Captioning}

For the purpose of image captioning, the image embeddings are passed to the cross-attention layers of the Image-grounded Text Decoder $D_\phi$, implemented as a Transformer Decoder~\cite{transformer}. This allows the model to generate contextually relevant descriptions based on visual input.

Given a collection of pairs of wound images and GPT-4o-generated text descriptions $\mathcal{D}_{GPT4} = \{(I_n, T_n)\}_{n=1}^N$, the BLIP model was fine-tuned by freezing the pre-trained Image Encoder and updating only the parameters $\phi$ of the Text Decoder. The objective is to predict the probability distribution of the next word in the sequence, given the input image and the previous words. The loss function associated with this task is the Language Modeling (LM) loss, which minimizes the negative log-likelihood of the text in an autoregressive manner. The LM loss function is expressed in Equation~\ref{eq:lm_loss}.

\begin{equation} L_{\text{LM}} = -\sum_{l=1}^{L}{\log{p(w_l,|,w_{<l}, I;, \phi)}} \label{eq:lm_loss} \end{equation}

Here, $p(w_l,|,w_{<l}, I;, \phi)$ represents the probability of the BLIP model outputting the correct $l$-th token $w_l$, given all previous tokens $w_{<l}$ in the textual sequence $T$ and the input image $I$. $L$ denotes the number of tokens in the text.

\subsubsection{Interpreting Captions with Image-Text Matching}

To interpret the generated captions on images, we use Image-Text Matching (ITM) and visualization techniques. The image embeddings and the generated descriptions are passed to the Image-grounded Text Encoder. We then apply Gradient-weighted Class Activation Mapping (Grad-CAM)~\cite{selvaraju2017grad} to the cross-attention layers of the Image-grounded Text Encoder to visualize the areas of the image that correspond to the textual descriptions.

Since the Image-grounded Text Encoder shares a similar architecture with the Image-grounded Text Decoder, we reused the fine-tuned cross-attention and feed-forward layers from the decoder in the encoder. However, it was still necessary to train the ITM head, which captures the fine-grained alignment between text and image. We employed the Binary Cross-Entropy (BCE) loss to predict whether the pairs of wound images and generated wound descriptions are matched.

\begin{figure}[!t]
\centering
  \includegraphics[width=0.98\textwidth]{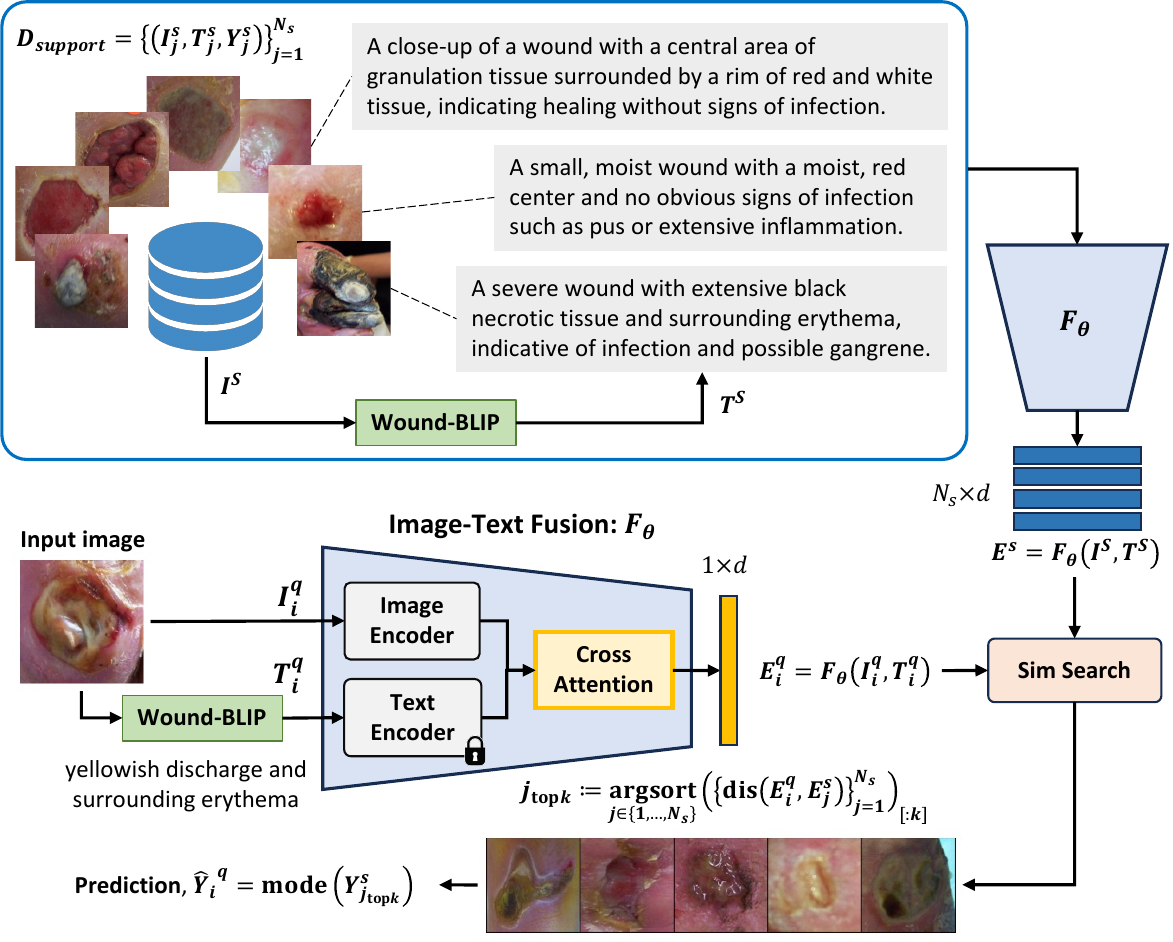}
  \caption{SCARWID Pipeline at Test Time: The infection classification starts by considering a query wound image $I^q_i$ as an input. After that Wound-BLIP generates a wound description $T^q_i$ corresponding to $I^q_i$. Then the Image-Text Fusion model, $F_\theta$, takes both $I^q$ and $T^q$ as inputs and transforms them into a $d$-dimensional multimodal embedding vector. Then the framework retrieves the top $k$-nearest neighbor objects in embedding spaces from the support document $\mathcal{D}_{support}$. Finally, the predicted status of the input $I^q_i$ is determined by the most common labels of $k$ retrieved objects $\text{mode}(Y^{s}_{j_{\text{top-k}}})$. Where, $j_{\text{top-k}}$ denotes top-$k$ indices.}
  \label{fig:rewilnet_pip}
  \vspace{-1.5mm}
\end{figure}

\subsection{SCARWID Model}
\label{sec:DB-MF}

By using the captions generated from our Wound-BLIP model as metadata, we integrated them with the corresponding wound images to predict infections in DFU images. This integration is performed by the \textbf{Image-Text Fusion} module $F_\theta$. Infection classification is then determined by retrieving the top-$K$ most similar instances from the support data collection $\mathcal{D}_{support}$ based on the fused image-text embeddings, as depicted in Fig.~\ref{fig:rewilnet_pip}.

\subsubsection{Image-Text Fusion Module}
This module consists of three components:

\underline{Image Encoder:} The DeiT (Data-efficient Image Transformers) model \cite{deit} is utilized to process the input image \(I\) and outputs an image embedding vector $E_I \in \mathbb{R}^{M \times d_i}$. Where $M$ is the number of patches in the image and $d_i$ is the embedding dimension.

\underline{Text Encoder:} The corresponding textual input \(T\) is processed by the CLIP-Text model \cite{CLIP}, which outputs a text embedding vector $E_T \in \mathbb{R}^{L \times d_t}$. Here, $L$ represents the number of tokens in the text, and $d_t$ is the embedding dimension.

\underline{Cross-Attention Layer:} To effectively fuse the information from both the image and text embeddings, a cross-attention mechanism is employed. This mechanism uses the image embedding as a query $Q$, with key $K$ and value $V$ derived from the text embedding. This structure allows the model to focus specifically on parts of the image relevant to the text description. The cross-attention layer's operation is expressed by Equation \ref{eq:cross_att}.

\begin{equation}
\text{Attention}(Q, K, V) = \text{softmax}\left(\frac{QK^T}{\sqrt{d}}\right)V
\label{eq:cross_att}
\end{equation}

Here, $Q = W^Q E_I$, $K = W^K E_T$, and $V = W^V E_T$, where $W^Q$, $W^K$, and $W^V$ are trainable parameters. The factor $\sqrt{d}$ serves as a scaling term to stabilize the gradients during training.

\subsubsection{Similarity-based Classification}

The final step in our classification process involves utilizing the cross-modal embedding $E = \text{Attention}(E_I, E_T, E_T)$ for classification. Rather than employing a traditional probability-based approach, our method treats an input image as a query image $I_i^q$ and its corresponding generated description from Wound-BLIP as query text $T_i^q$ that are then both fed into the Image-Text Fusion module, producing the query embedding $E_i^q = F_\theta(I_i^q, T_i^q)$. 

Next, we search for the top $k$ similar pairs from a labeled support document $\mathcal{D}_{support} = \{(I^s_j, T^s_j, Y^s_j)\}_{j=1}^{N_s}$. Here, $I^s_j$, $T^s_j$, and $Y^s_j$ represent the support images, corresponding Wound-BLIP generated texts, and their respective labels, with $N_s$ denoting the total number of items in $\mathcal{D}_{support}$. Each support item's cross-modal embedding is computed as $E^s_j = F_\theta(I_j^s, T_j^s), \ \forall j \in \{1, ..., N_s\}$.

The predicted label $\hat{Y_i}^q$ for the query image $I_i^q$ is determined by identifying the most common label among the top-$k$ objects, based on the minimum Euclidean distance in embedding space, calculated as $\text{dis}(E^q_i, E^s_j) = \sqrt{(E_i^q - E^s_j)^2}$. The set of indices for the top-$k$ similar objects is denoted by $j_{\text{top-k}}$, and formally, the label determination process can be described as follows:
\begin{equation}
j_{\text{top-k}} = \mathrm{argsort}\Bigl(\{\mathrm{dis}(E_i^q, E_j^s) : j = 1,\ldots,N_s\}\Bigr)_{[:k]}
\label{eq:top-k}
\end{equation}
\begin{equation}
\hat{Y}_i^q = \text{mode}(Y^s_{j_{\text{top-k}}})
\end{equation}

\subsubsection{Learning Similarity using a Triplet Loss Function} To learn the similarity between objects in the cross-modal embedding space, we leveraged the triplet loss function~\cite{facenet} for optimizing parameters $\theta$ of our Image-Text Fusion module $F_\theta$. This works by minimizing the distance between an anchor object $x^{(a)}$ and a positive object $x^{(p)}$ with the same identity while maximizing the distance between the anchor object and a negative object $x^{(n)}$ with a different identity. Here $x^{(*)}$ is denoted as a pair of $($wound image $I_j$, text description $T_j$ $)$.

\begin{equation}
\begin{aligned}
L_{triplet} = \mathbb{E}\bigg[ & \left(\|F_\theta(x^{(a)}) - F_\theta(x^{(p)})\|^2_2  - \|F_\theta(x^{(a)}) - F_\theta(x^{(n)})\|^2_2 + \alpha \right)_+ \bigg]
\end{aligned}
\label{eq:triplet_loss}
\end{equation}

The margin $\alpha$ is set to 1, indicating the desired separation between similar and dissimilar pairs.

\subsection{Dataset Preparation and Processing}
\label{sec:dataset}

\subsubsection{DFU Infection Dataset}
The DFU Infection Dataset is derived from the \textbf{Part-B DFU Dataset} \cite{partb_DFU}, which encompasses two categories of DFU diseases: ischemia and infection. This data set was compiled from patient wound images obtained at the Lancashire Teaching Hospital with permission for research granted by the UK National Health Service (NHS). The images were labeled by two healthcare professionals, consultant physicians specializing in diabetic foot conditions, based solely on visual assessments without referencing medical notes or clinical tests. This project focuses on infection classification based on the visual appearance of an image.

The available DFU infection dataset used in this project contains regions of interest for infection classification, which consists of 2,946 natural augmented patches with infection and 2,946 natural augmented patches of non-infection where the natural data augmentation is capturing multiple magnifications of the same wound image. Each DFU patch measures $224 \times 224 \times 3$ pixels.

\underline{Data Pre-processing}:
To prevent data leakage, we partitioned the dataset on a \textit{subject-wise basis}, ensuring that all images from a given subject were included in either the training, validation, or test sets. The data was split into training (60\%), validation (20\%), and testing (20\%) sets. Five-fold cross-validation was employed to evaluate model performance across different test partitions.

\subsubsection{Metadata Generation with GPT-4o}

\begin{figure}[thb]
  \centering
  \includegraphics[width=1 \linewidth]{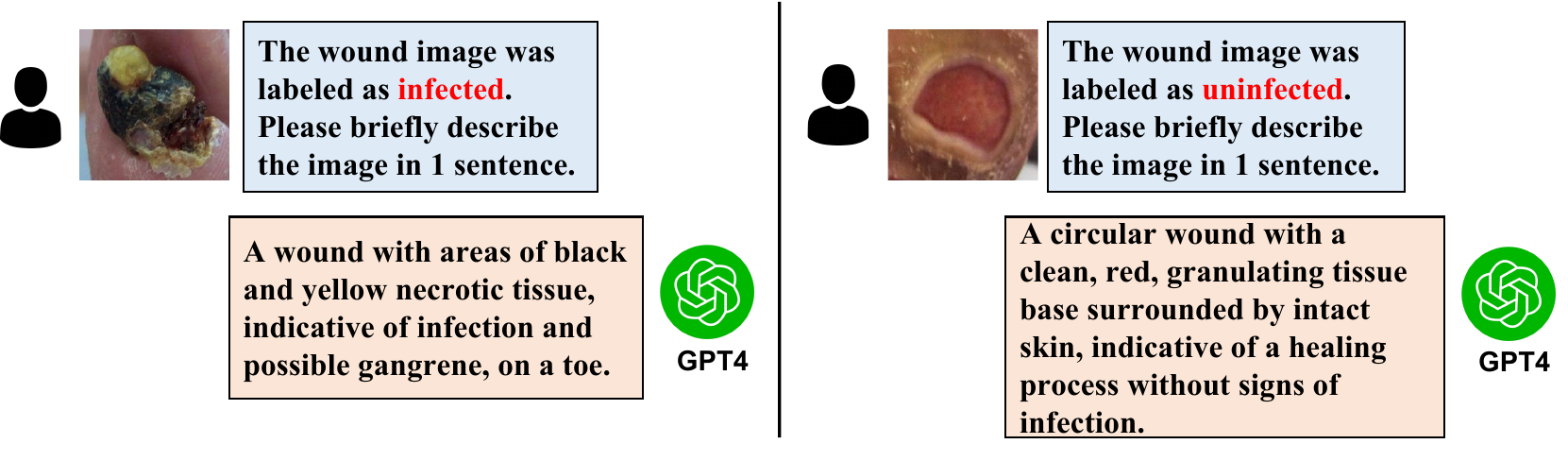}
  \caption{Label-Guided Prompting for Textual Metadata Generation:
\{\textit{System}: You are a wound care physician\}.
A user's prompt is as follows. 
\{\textit{User}: $<$\textbf{image}$>$ | The wound image was labeled as $<$\textbf{label}$>$. Please briefly describe the image in 1 sentence.\}
}
  \label{fig:prompt-guided}
\end{figure}

To address the significant challenges of predicting infection only by the appearance of the wound in an image described in Sec.~\ref{sec:introduction}, we utilize GPT-4o (i.e., gpt-4o-2024-08-06 version) with our label-guided prompting technique to generate textual descriptions corresponding to each wound image. This technique involves initially informing the model of the ground-truth infection label assigned by wound specialists. Subsequently, GPT-4o is prompted to identify and describe characteristics that potentially influenced the specialists’ diagnostic decisions. This process is illustrated in Fig.~\ref{fig:prompt-guided}.

For this study, we randomly selected 500 images belonging to each of the target infected and uninfected classes, resulting in a total of 1,000 images in the training set . These images were then processed using GPT-4o to generate textual descriptions of the wound image to generate collection pairs of images and texts: $\mathcal{D}_{GPT4} = \{(I_n, T_n)\}_{n=1}^N$. Table~\ref{tab:gpt4-generated} presents examples of infected and uninfected wounds and the corresponding descriptions generated by GPT-4o.

\begin{table}[!ht]
\centering
\caption{Examples of Wound Descriptions generated by GPT-4o using Label-Guided Prompting of Images along with ground truth Infection Labels}
\label{tab:gpt4-generated}

\renewcommand{\arraystretch}{1} 

\begin{tabular}{|c|c|c|}
\hline
\textbf{Infection} & \textbf{Image} & \textbf{GPT-4o Description} \\
\hline

No 
& \raisebox{-0.5\height}{\includegraphics[width=0.12\linewidth]{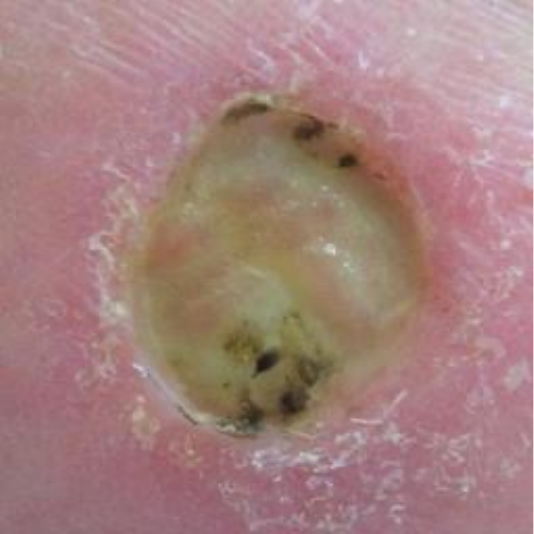}}
& \shortstack{A shallow wound with a moist, yellowish base and some\\ black areas, surrounded by healthy pink skin.} \\

\hline

No
& \raisebox{-0.5\height}{\includegraphics[width=0.12\linewidth]{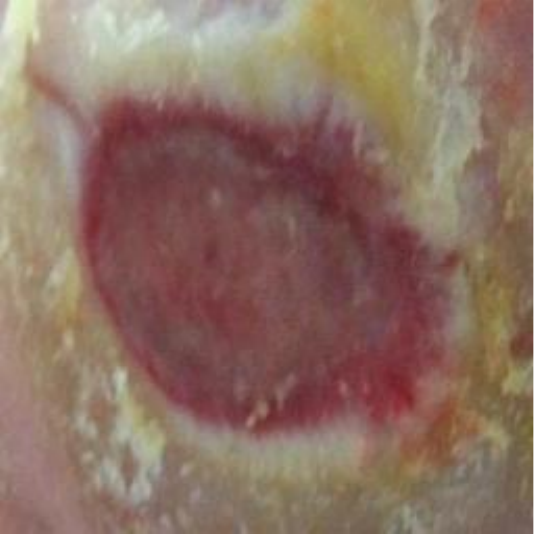}}
& \shortstack{A circular wound with a red, granulating tissue bed \\ surrounded by yellowish slough and erythematous skin, \\ indicating a healing stage.} \\

\hline

Yes
& \raisebox{-0.5\height}{\includegraphics[width=0.12\linewidth]{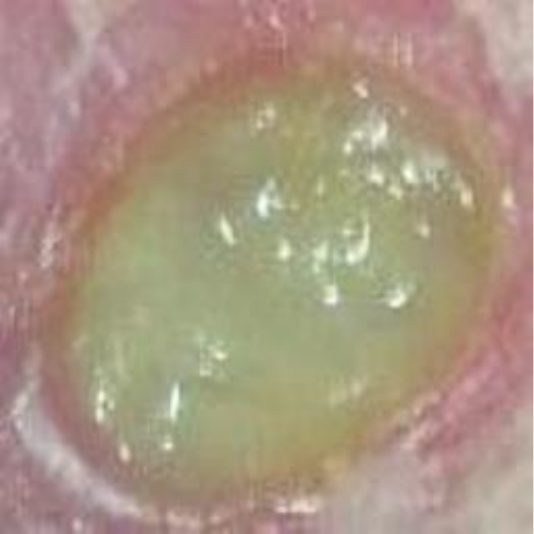}}
& \shortstack{A wound with yellowish exudate and reddened edges, \\suggesting signs of infection.} \\

\hline

Yes
& \raisebox{-0.5\height}{\includegraphics[width=0.12\linewidth]{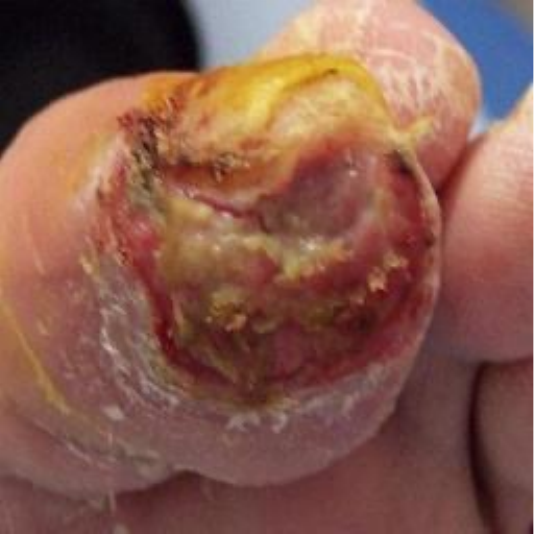}}
& \shortstack{A close-up of a wound on a toe, characterized by redness,\\swelling, and yellowish discharge, indicative of infection.} \\

\hline
\end{tabular}

\end{table}

\subsubsection{Synthetic Image Augmentation}
\label{sec:augmentation}
Diffusion models~\cite{DDPM, DDIM}, a novel class of generative models, utilize diffusion processes to generate high-quality images by progressively reducing noise in multiple iterations. Recent studies have utilized diffusion models for image augmentation~\cite{diff_aug, diff_aug2, diff_aug3}, significantly enhancing the accuracy of baseline deep learning models in image classification tasks, including medical imaging analysis.

\begin{figure}[!ht]
  \centering
  \begin{subfigure}[b]{0.49\textwidth}
    \centering
    \includegraphics[width=\linewidth]{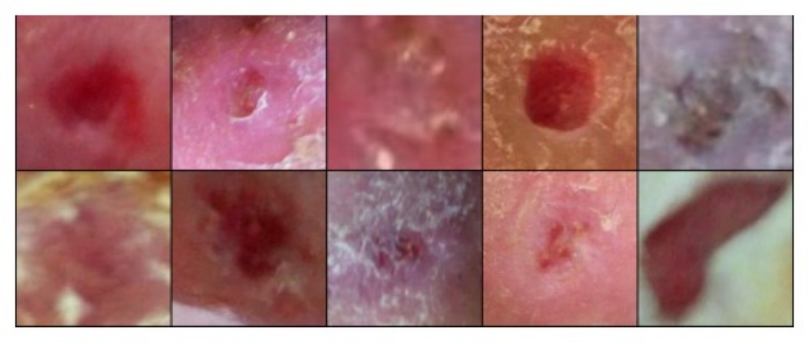}
    \caption{Synthesized Images (No Infection)}
  \end{subfigure}
\hfill
  \begin{subfigure}[b]{0.49\textwidth}
    \centering
    \includegraphics[width=\linewidth]{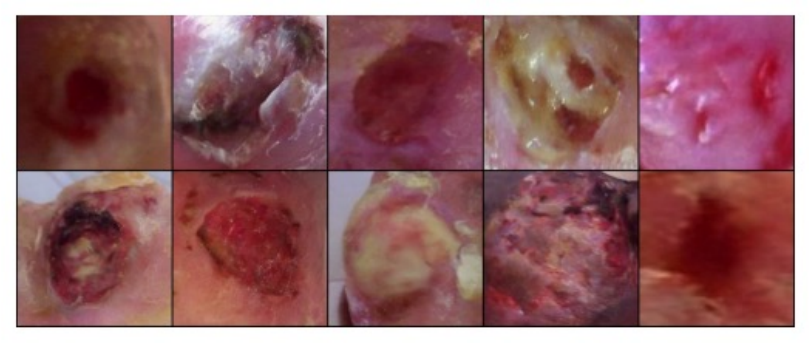}
    \caption{Synthesized Images (Infection)}
  \end{subfigure}

  \caption{Examples of Conditional Synthesized Wound Images by the Diffusion Model}
  \label{fig:gen_images}
\end{figure}

In this study, the label-conditional latent diffusion model used in ConDiff~\cite{ConDiff} was used to generate wound images that were conditioned on infection status (each of 1200 images). These 2400 generated images were added to the training data as augmented images.
Classifier-free guidance with the DDIM sampling process~\cite{CFG} was used to synthesize images of size $256\times256$. Examples of conditional synthesized images are shown in Fig.~\ref{fig:gen_images}.
The guidance scale and the sampling steps were set to 1.5 and 30 respectively. To prevent data leakage when evaluating classification models on the testing partitions, this diffusion model was only employed on DFU images in the training partition.

\section{Experimental Results}
\label{sec:experiments}

\subsection{Experimental Setup}
\label{sec:setup}
\subsubsection{\textbf{Wound-BLIP model's Fine-tuning configuration}}

To fine-tune the Wound-BLIP model, the pretrained parameters of \textit{blip-image-captioning-base} \footnote{\url{https://huggingface.co/Salesforce/blip-image-captioning-base}} were utilized. The model was optimized on pairs of metadata $\mathcal{D}_{GPT4}$, treating images as inputs and texts as outputs. The objective was to minimize the LM loss function in Equation~\ref{eq:lm_loss}. Training was done for 20 epochs using the AdamW optimizer with a learning rate of $1 \times 10^{-5}$.

\subsubsection{\textbf{SCARWID model's training configuration}}

The SCARWID model's configuration consists of the following three modules:

\begin{itemize}
\item \textbf{Image Encoder parameters} were initialized using the \textit{deit-base-distilled-patch16-224} model%
\footnote{\url{https://huggingface.co/facebook/deit-base-distilled-patch16-224}}.
\item \textbf{Text Encoder parameters} were initialized with the CLIP-Text encoder from the \textit{clip-vit-large-patch14} model%
\footnote{\url{https://huggingface.co/openai/clip-vit-large-patch14}}.
\item \textbf{Cross-Attention Layer hyperparameters} were set to 2 attention heads, an embedding dimension of 768 (matching the output sizes of both Image and Text Encoders), and a projection dimension (i.e., cross-modal embedding) of 256.
\end{itemize}

As previously mentioned in Sec.~\ref{sec:dataset}, 5-fold cross-validation was employed for training and testing deep learning models. For our SCARWID model, 
image descriptions generated by Wound-BLIP were paired with their corresponding wound images as input. The model was trained for 30 epochs using the AdamW optimizer, with a learning rate of $1 \times 10^{-4}$, with the goal of minimizing the triplet loss function defined in Equation ~\ref{eq:triplet_loss}.  

During inference, labeled support data $\mathcal{D}_{support}$ were randomly sampled from training images, ensuring that at most one image from each subject was selected. The total number of samples in $\mathcal{D}_{support}$ was set to 1024. These data were stored as embedding vectors $E^s$. When predicting a new input query image $I_i^q$, SCARWID computes a cross-modal embedding vector $E_i^q$ from $I_i^q$ and its associated caption $T_i^q$. To determine the label of a given query image, the labels of the top-5 similar objects from $\mathcal{D}_{support}$ were retrieved. 

Experiments were done in Python 3.9 using the following software libraries: PyTorch~1.13.1, \texttt{torchvision} 0.14.1, \texttt{transformers} 4.42.4, and \texttt{salesforce-lavis} 1.0.2. An NVIDIA A100 GPU was used to train the models.

\subsection{Evaluation Metrics}

To evaluate our proposed framework for DFU infection classification task, the following metrics are considered. 

\begin{itemize}
\item \textbf{Accuracy} $ACC = \frac{TP+TN}{P+N}$, where $TP$ is the number of true positive predictions, $TN$ is the number of true negative predictions, $P$ is the positive label (infected), and $N$ is the negative label (not infected).
\item \textbf{Sensitivity (SEN)} or recall reflects the proportion of actual positives that are correctly identified: $SEN = \frac{TP}{TP+FN}$, where $FN$ denotes the number of false negative predictions.
\item \textbf{Specificity (SPC)} reflects the proportion of actual negatives that are correctly identified: $SPC = \frac{TN}{TN+FP}$, where $FP$ denotes the number of false positive predictions.
\item \textbf{Positive Predictive Value (PPV)} or precision is the proportion of positive predictions that are true positives. $PPV = \frac{TP}{TP+FP}$.
\item \textbf{F1-score} is the Harmonic Mean of Precision and Recall: $F1 = 2 \cdot \frac{PPV \cdot SEN}{PPV + SEN}$. 
\end{itemize}


%
\subsection{SOTA Baseline Models}
\label{sec:baseline}

Recent deep-learning architectures were selected as baselines for wound infection classification from images. These include custom CNN architectures such as CNN-Ensemble~\cite{partb_DFU} and DFU-RGB-TEX-Net~\cite{al2022diabetic}. Additionally, ConDiff~\cite{ConDiff}, a distance-based generative discrimination model, was also selected. EfficientNet~\cite{efficientnet} was chosen as it was the most effective CNN-based model for the detection of infections from wound images~\cite{yap2021analysis}. Transformer-based models such as ViT~\cite{vit}, DeiT~\cite{deit}, SwinV2~\cite{swinv2}, and EfficientFormer~\cite{efficientformer}, which have demonstrated superior performance over traditional CNN-based models in wound infection classification~\cite{galdran2021convolutional, ConDiff}, were also included.

\subsection{Deep Learning for Image Classification with Image Augmentations}
\label{sec:Aug_compare}

\begin{table}[!ht]
\centering 
\small 
\caption{Quantitative comparison of infection classification on test images with different data augmentation techniques for training deep learning models. Values are the mean (stdev) obtained from optimal models over 5-fold cross-validation. Bold values indicate the highest scores.}  
\label{tab:Aug_compare} 

\begin{tabular}{@{}ccccccc@{}} 
\hline
Model              & Augmentation & Accuracy              & Sensitivity            & Specificity           & PPV                   & F1-score              \\ \hline
EfficientNet-B0    & Manual       & 0.734 (0.022)         & \textbf{0.760 (0.032)}         & 0.708 (0.063)         & 0.727 (0.037)         & 0.741 (0.014)         \\
                   & Diffusion    & \textbf{0.762 (0.013)} & 0.747 (0.038)         & \textbf{0.778 (0.043)} & \textbf{0.774 (0.026)} & \textbf{0.759 (0.018)} \\ \hline
ViT-Base           & Manual       & 0.728 (0.015)         & \textbf{0.721 (0.044)}         & 0.735 (0.027)         & 0.734 (0.012)         & 0.726 (0.022)         \\
                   & Diffusion    & \textbf{0.752 (0.010)} & 0.699 (0.044)         & \textbf{0.805 (0.040)} & \textbf{0.785 (0.022)} & \textbf{0.738 (0.014)} \\ \hline
DeiT-Base          & Manual       & 0.745 (0.011)         & 0.774 (0.054)         & 0.716 (0.067)         & 0.737 (0.034)         & 0.753 (0.013)         \\
                   & Diffusion    & \textbf{0.773 (0.009)} & \textbf{0.782 (0.071)} & \textbf{0.763 (0.082)} & \textbf{0.775 (0.044)} & \textbf{0.775 (0.015)} \\ \hline
SwinV2-Tiny        & Manual       & 0.737 (0.015)         & 0.749 (0.024)         & 0.725 (0.040)         & 0.735 (0.024)         & 0.741 (0.014)         \\
                   & Diffusion    & \textbf{0.770 (0.018)} & \textbf{0.787 (0.058)} & \textbf{0.752 (0.090)} & \textbf{0.769 (0.050)} & \textbf{0.774 (0.014)} \\ \hline
EfficientFormer-L1 & Manual       & 0.735 (0.022)         & 0.763 (0.046)         & 0.706 (0.045)         & 0.725 (0.024)         & 0.743 (0.014)         \\
                   & Diffusion    & \textbf{0.766 (0.014)} & \textbf{0.764 (0.045)} & \textbf{0.768 (0.046)} & \textbf{0.772 (0.027)} & \textbf{0.767 (0.017)} \\ \hline
\end{tabular}
\end{table}

We trained SOTA image classification models using two different data augmentation techniques: (1) \textit{Traditional image augmentation operations}, which included random crops, vertical and horizontal flips, rotations and adjustments to brightness, contrast and saturation; and (2) \textit{Synthetic Augmentation}, utilizing images generated by a diffusion model (see Sec.~\ref{sec:augmentation}).

As shown in Table~\ref{tab:Aug_compare}, the inclusion of synthetic images from the diffusion model substantially improves the performance of SOTA deep learning models across most metrics, increasing accuracy by 2.5-4.5\% for infection classifications from DFU images. In particular, transformer-based models such as DeiT-Base and SwinV2-Tiny achieved enhanced performance with synthetic augmentation compared to EfficientNet-B0, likely due to the increased variety of images.

\subsection{Performance Comparison of SCARWID with SOTA baselines}
\label{sec:Performance}

\begin{table*}[!ht]

\centering 
\caption{Comparison of infection classification performance of  SOTA baseline models on test partitions. Values are the mean (stdev) obtained from the best performing over 5-folds. Bold values indicate the highest scores.}  
\label{tab:SOTA_comparison} 
\small
\resizebox{\textwidth}{!}{
\begin{tabular}{clccccc}
\hline
\multicolumn{2}{c}{Model w/ Augmentation Technique}  & Accuracy & Sensitivity & Specificity & PPV & F1-score \\ \hline
\multirow{4}{*}{\begin{tabular}[c]{@{}c@{}}Probability \\ based\end{tabular}}
& Ensemble CNN \cite{partb_DFU} & 0.727 (0.025) & 0.709 (0.044) & 0.744 (0.050) & 0.735 (0.036) & 0.722 (0.028) \\
& DFU-RGB-TEX-Net \cite{al2022diabetic} & 0.742 (0.018) & 0.751 (0.063) & 0.734 (0.050) & 0.741 (0.021) & 0.744 (0.036) \\
& DeiT-Base w/ Manual Aug  & 0.745 (0.011) & 0.774 (0.054) & 0.716 (0.067) & 0.737 (0.034) & 0.753 (0.013) \\
& DeiT-Base w/ Diffusion Aug (Ours) & 0.773 (0.009) & 0.782 (0.071) & 0.763 (0.082) & 0.775 (0.044) & 0.775 (0.014) \\ \hline
\multirow{4}{*}{\begin{tabular}[c]{@{}c@{}}Similarity \\ Based\end{tabular}} 
& ConDiff $^a$ \cite{ConDiff}    & 0.780 (0.024) & 0.817 (0.033) & 0.743 (0.033) & 0.763 (0.023) & 0.788 (0.023) \\
& SCARWID (Text Only, Ours)   & 0.750 (0.014) & 0.783 (0.017) & 0.716 (0.023) & 0.736 (0.015) & 0.759 (0.012) \\
& SCARWID (Image Only, Ours)  & 0.784 (0.022) & 0.831 (0.023) & 0.736 (0.034) & 0.761 (0.023) & 0.795 (0.019) \\
& \textbf{SCARWID (Image \& Text, Ours)} $^b$  & \textbf{0.814} (0.011) & \textbf{0.852} (0.024) & \textbf{0.777} (0.011) & \textbf{0.790} (0.006) & \textbf{0.820} (0.010) \\ \hline        
\end{tabular}}
\caption*{\footnotesize{
\textit{$^a$ ConDiff was retrained and evaluated on the same training and testing partitions of the Part-B DFU dataset as SCARWID.\\
$^b$ Throughout the paper, we refer to SCARWID (Image \& Text) simply as SCARWID. \\
* Diffusion image augmentations were applied for training SCARWID models.}}}
\end{table*}

Building on insights from experiments detailed in Sec.~\ref{sec:Aug_compare}, which highlighted the effectiveness of synthetic augmentation, we further incorporated diffusion-generated images and their descriptions from Wound-BLIP into the training process of SCARWID.

As detailed in Table~\ref{tab:SOTA_comparison}, SCARWID (Image \& Text) demonstrates superior performance, achieving an average accuracy of approximately 81.4\% and an average F1-score of 82.0\%, which significantly outperforms baselines. Furthermore, SCARWID exhibits lower standard deviations in evaluation scores across 5 folds during cross-validation, highlighting its robustness, especially when compared to probability-based models. In clinical scenarios, the highest sensitivity achieved by SCARWID (85.2\%) is particularly valuable in the context of wound care management, as it improves the model’s ability to detect infections early, enabling caregivers to flag and examine potentially infected wounds more closely, and administer antibiotic treatment or surgical procedures to reduce severe complications such as amputation. Additionally, SCARWID's good specificity score reduces unnecessary referrals, allowing better resource-utilization in clinics.

Further insights into the utility of generating corresponding wound descriptions are gained by comparing SCARWID (Image \& Text) with the SCARWID (Image Only). As shown in Table~\ref{tab:SOTA_comparison}, without the support of the wound descriptions generated, SCARWID (Image Only) achieves a sensitivity of 83.1\%, about 2\% lower than that of SCARWID (Image \& Text) and a specificity score of 73.6\% is 4\% lower than that of SCARWID (Image \& Text). In addition, we observe that the standard deviations of its evaluation scores are higher than those of SCARWID (Image \& Text). This result suggests that the inclusion of the Wound-BLIP generated descriptions helps improve the model robustness and generalization of SCARWID, and mitigates the fine-grained appearance challenge with high inter-class similarity by providing textual context that distinctly characterizes wound attributes, enabling more accurate classification.

Likewise, classifying wound infections using only the generated text descriptions also underperforms combining them with the wound image. SCARWID (Text Only) even achieved lower sensitivity (78.3\%) and specificity (71.6\%) scores than SCARWID (Image Only). This result underscores the limitations of the Wound-BLIP model in generating accurate and reliable wound descriptions on its own, which might lead to less precise or even erroneous diagnoses when used without concurrent image analysis. 

\subsection{SCARWID Explainability}
\label{sec:explain}

\subsubsection{Visualization of Cross-modal Embedding} 
\begin{figure*}[!t]
\centering
  \includegraphics[width=1\textwidth]{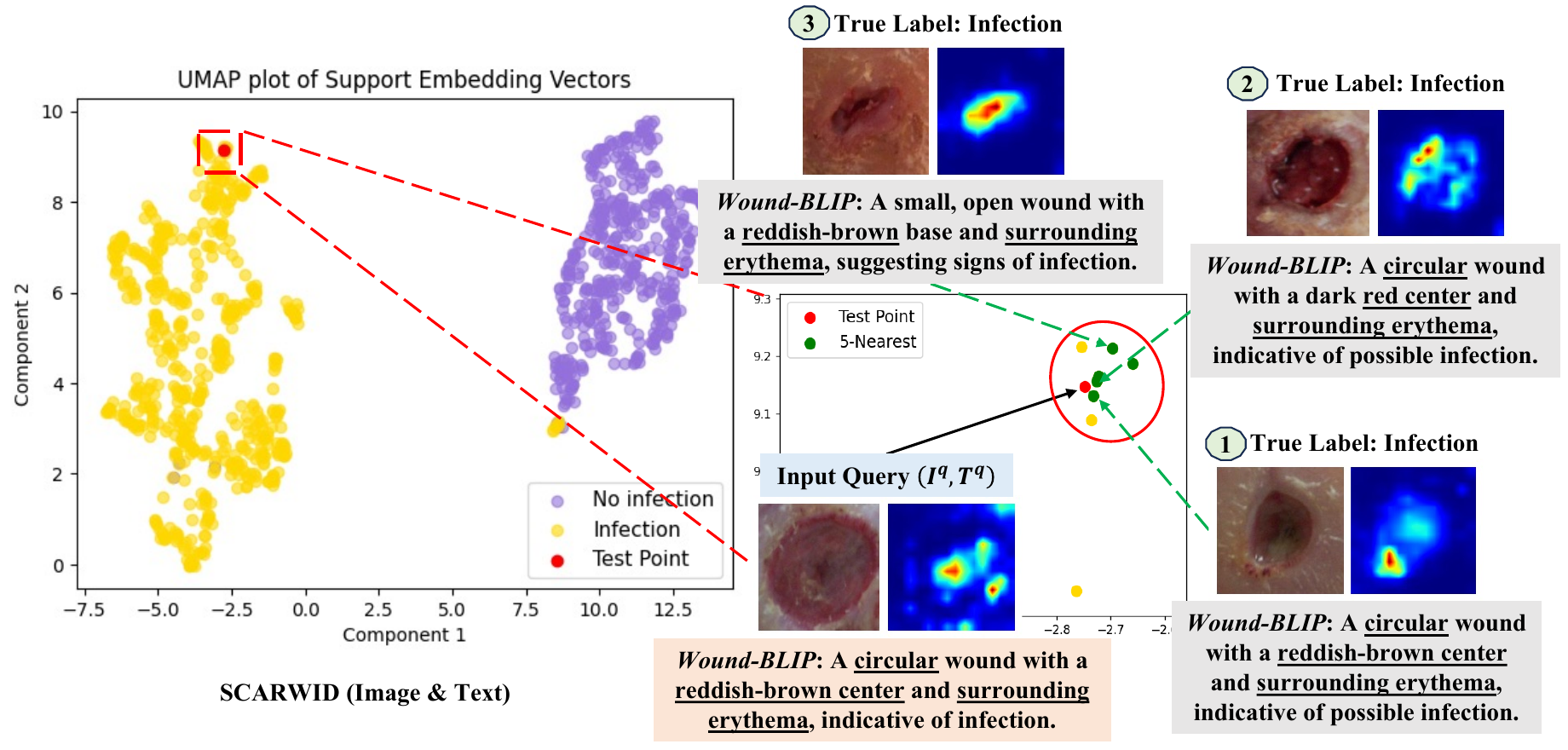}
  \caption{(1) UMAP plot of support cross-modal embedding computed by the Image-Text Fusion module $F_\theta$ of the SCARWID framework and (2) visualization of test image prediction with its $k$-nearest pairs of support wound images and their corresponding generated description. The corresponding attention heatmap is shown on the right side of each image.}
  \label{fig:SCARWIDNet_vis}
\end{figure*}

The  plot in Fig.~\ref{fig:SCARWIDNet_vis} shows cross-modal embedding vectors between image-text pairs $(I^s, T^s) \in \mathcal{D}_{support}$. It is observed that infected and uninfected wounds are separated into two distinct clusters. 

\subsubsection{Attention Map Visualization with Attention Rollout}
Attention Rollout~\cite{attention_roll} is a method employed to visualize and elucidate which parts of an input image are predominantly focused on by a Vision Transformer-based model during its decision-making process. This technique involves the aggregation of attention weights from all attention heads across all layers of a transformer, thereby illustrating the areas deemed most predictive by the model.


In Fig.~\ref{fig:SCARWIDNet_vis}, we illustrate an example of SCARWID predicting an infected test image, depicted as a red circle within the UMAP plot. This test point is encircled by infected wounds, marked by yellow points, demonstrating the model's effectiveness in clustering similar cases. On the right panel of Fig.~\ref{fig:SCARWIDNet_vis}, a zoomed-in view of the area around the red dot shows its 3-Nearest Neighbors, which helps elucidate the context of its classification. It is important to note that while SCARWID typically considers the labels of the top-5 most similar objects during its decision-making process, to improve the clarity of the visualization, this example focuses on only the three closest objects. The following observations can be made:
\begin{itemize}
\item \underline{Top-$k$ retrieval support pairs}: The descriptions generated for the wound images, numbered 1, 2, and 3, share meaningful similarities with the query's generated description, emphasizing key phrases highlighted in red. This similarity indicates that our $F_\theta$ effectively synergizes information from both modalities. Although relying solely on textual information does not achieve high accuracy (as seen with CLIPText in Table~\ref{tab:SOTA_comparison}), the fusion of visual and textual data leads to more robust decision-making.

\item \underline{Rollout attention heatmaps}: The image encoder's attention maps reveal focus areas in red, corresponding closely to the text descriptions. For instance, the red spot in the center of the the input query's attention map  (bottom right image of Fig.~\ref{fig:SCARWIDNet_vis}) matches the \textit{reddish-brown center} noted in the description. Additionally, the model's focus extends to the wound's edges, aligning with the mention of \textit{surrounding erythema}. This correlation underscores the cross-attention layer's ability to effectively integrate information from both images and text.
\end{itemize}

\begin{figure*}[!htbp]
\centering
  \includegraphics[width=0.9\textwidth]{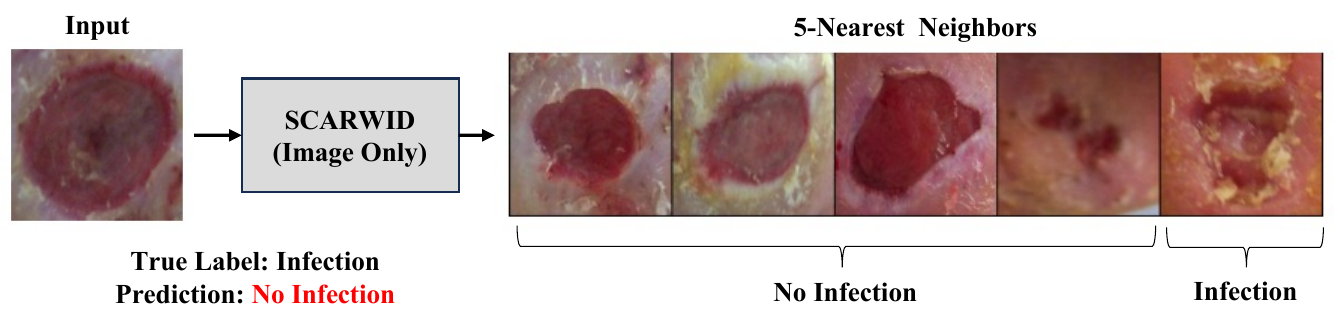}
  \caption{Example of the misclassified SCARWID (Image Only) prediction from an input image in Fig.~\ref{fig:SCARWIDNet_vis}, and its 5 similar images retrieved from the support document.}
  \label{fig:SCARWID_image_vis}
\end{figure*}

\begin{figure*}[!thbp]
    \centering
    
    \begin{subfigure}{\linewidth}
        \centering
        \includegraphics[width=0.9\textwidth]{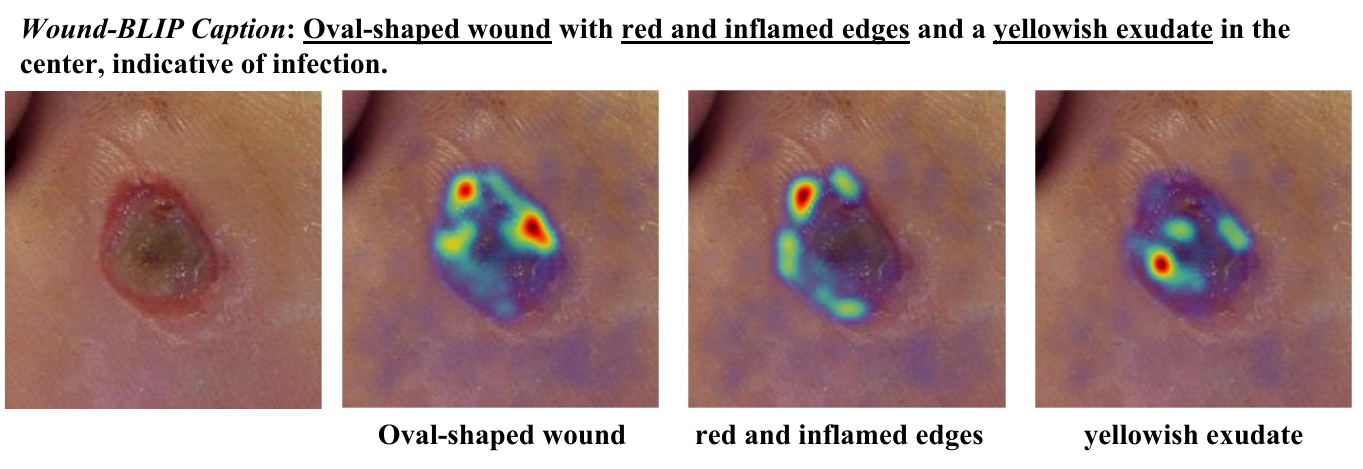} 
        \caption{Grad-CAM visualization of infected wound 1}
        \label{fig:sub1}
    \end{subfigure}

    \begin{subfigure}{\linewidth}
        \centering
        \includegraphics[width=0.9\textwidth]{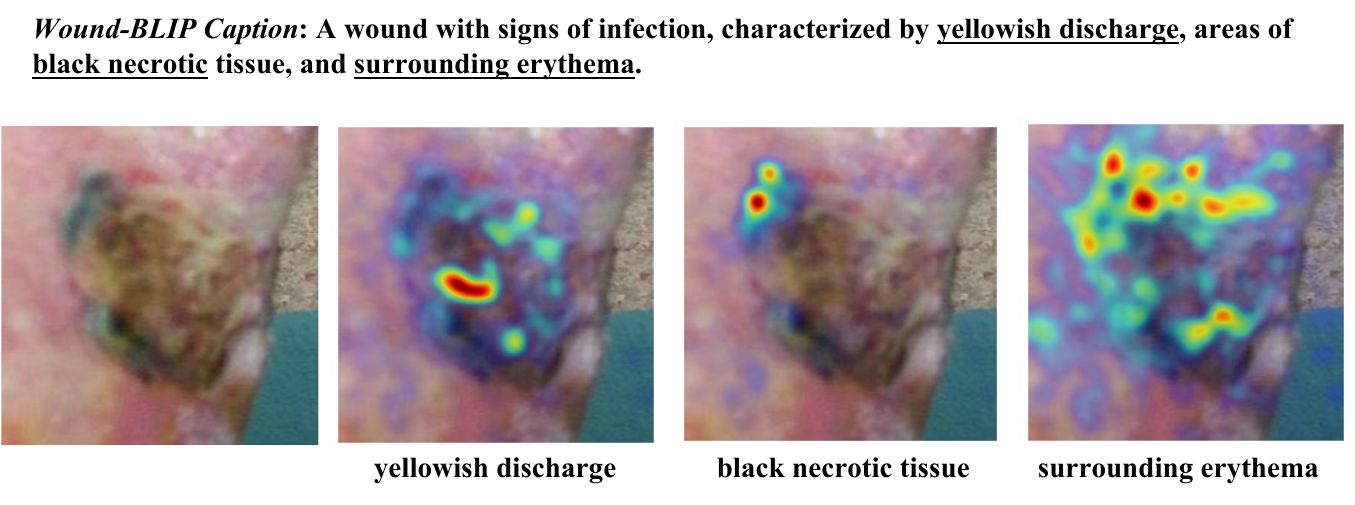}
        \caption{Grad-CAM visualization of infected wound 2}
        \label{fig:sub2}
    \end{subfigure}

    \caption{Grad-CAM visualizations illustrating the localization of wound regions corresponding to the Wound-BLIP-generated wound descriptions. The heatmaps highlight key areas, facilitating image-text matching for enhanced wound interpretation.}
    \label{fig:gradcam_vis}
\end{figure*}

Next, we consider the same test image from Fig.~\ref{fig:SCARWIDNet_vis} but without including generated text. As depicted in Fig.~\ref{fig:SCARWID_image_vis},  four out of the top five similar support images identified by SCARWID (Image Only) are labeled as no infection, contradicting the ground truth label of the test image. This scenario underscores the SCARWID (Image Only)'s ongoing struggle with interclass similarity issues, evidenced by the fact that the basic wound characteristics from support images, such as the red circular wounds of the five nearest neighbors, closely resemble those of the query image. This highlights the challenge of achieving accurate classifications based solely on visual features without the contextual support of generated textual descriptions.

\subsubsection{Wound-BLIP Caption Interpretability with Grad-CAM}

Fig.~\ref{fig:gradcam_vis} illustrates text localization on wound images, showcasing the ability of Wound-BLIP to generate meaningful captions and localize important wound features via Grad-CAM visualizations. In Fig.~\ref{fig:sub1}, Wound-BLIP generates a caption for an infected wound with the descriptors \textit{red and inflamed edges} and \textit{yellowish exudate}. The Grad-CAM visualization focuses precisely on the wound's edges, where redness and inflammation are prominent, aligning well with the clinical signs of infection. Additionally, the visualization highlights the yellow watery area of the wound, consistent with the \textit{yellowish exudate} description, a common feature of infected wounds.

\begin{figure*}[!thbp]
    \centering
    \begin{subfigure}{\linewidth}
        \centering
        \includegraphics[width=0.9\textwidth]{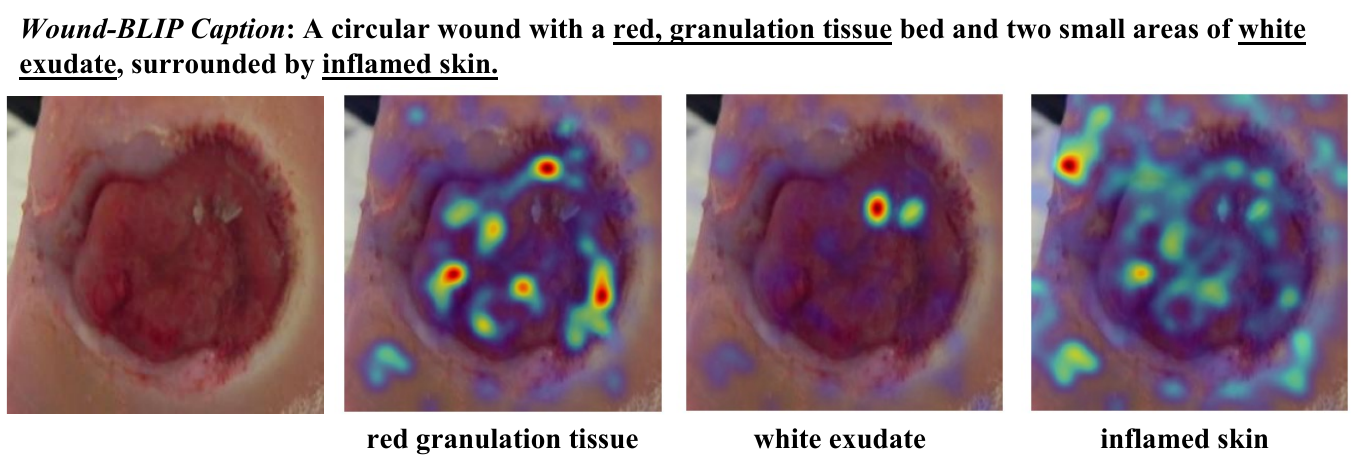}
        \caption{Grad-CAM visualization of uninfected wound 1}
        \label{fig:sub3}
    \end{subfigure}

    \begin{minipage}{0.9\linewidth}
        \centering
        \begin{subfigure}{0.48\linewidth}
            \centering
            \includegraphics[width=\linewidth]{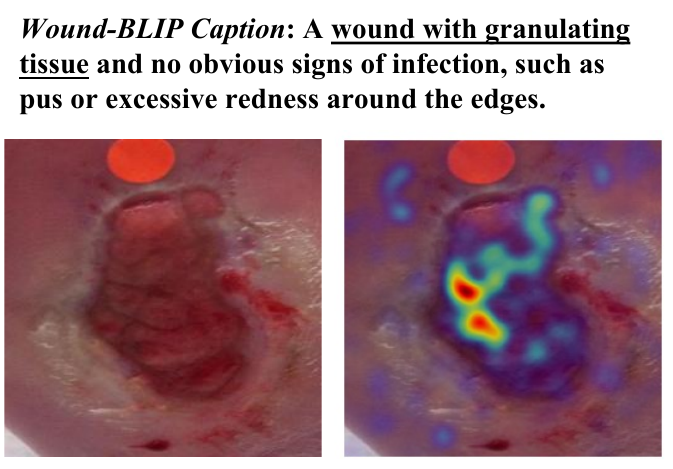}
            \caption{Grad-CAM visualization of uninfected wound 2}
            \label{fig:sub4}
        \end{subfigure}
        \hfill
        \begin{subfigure}{0.48\linewidth}
            \centering
            \includegraphics[width=\linewidth]{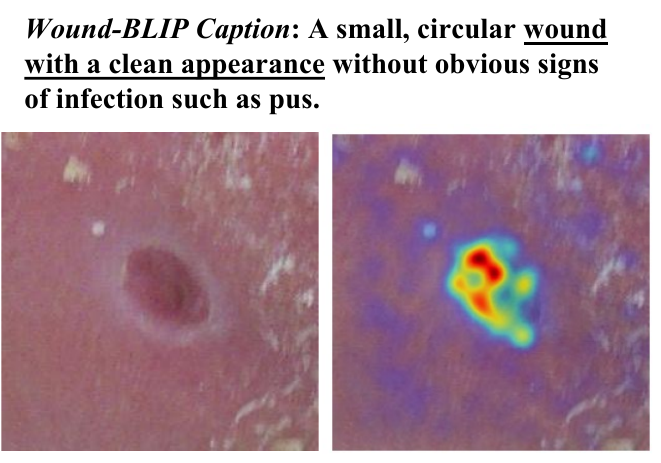}
            \caption{Grad-CAM visualization of uninfected wound 3}
            \label{fig:sub5}
        \end{subfigure}
    \end{minipage}

    \caption{Grad-CAM visualizations illustrating the localization of wound regions corresponding to the Wound-BLIP-generated wound descriptions. The heatmaps highlight key areas, facilitating image-text matching for enhanced wound interpretation.}
    \label{fig:gradcam_vis2}
\end{figure*}

Similarly, Fig.~\ref{fig:sub2} shows an infected wound, where Wound-BLIP identifies the feature \textit{yellowish discharge}. The Grad-CAM heatmap highlights lighter, moist, yellow areas toward the upper right of the wound (in light yellow). The red region of the heatmap indicates the thick, yellow fibrous region that aligns with the descriptive term yellowish color of the discharge. For the wound feature \textit{black necrotic tissue}, the Grad-CAM heatmap specifically focuses on the black area at the top of the wound, suggesting necrotic tissue. The heatmap also highlights redness in the surrounding skin, indicating potential \textit{surrounding erythema}, although some highlighted regions may be slightly off, focusing on the bottom-left portion of the wound. This slight discrepancy may arise because that particular area appears comparatively redder than other parts of the wound.

In contrast, Fig.~\ref{fig:sub3} and Fig.~\ref{fig:sub4} depict uninfected wounds, where the captions emphasize healthy granulation tissue. The Grad-CAM visualizations focus on areas with soft, red, and moist tissue, a characteristic of healthy wound healing. This demonstrates the model’s ability to distinguish between infected and uninfected tissues. For example, in Fig.~\ref{fig:sub3}, Wound-BLIP successfully localizes the small regions of \textit{white exudate}, while also highlighting the surrounding skin described as \textit{inflamed skin}.

Finally, Fig.~\ref{fig:sub5} presents an uninfected wound with a clear margin and no obvious signs of infection, as indicated by the text, \textit{wound with a clean appearance}. The Grad-CAM visualization highlights the central region of the wound, focusing on the healthy tissue.

These examples highlight the capability of Wound-BLIP to localize relevant clinical features of both infected and uninfected wounds, matching them to the text descriptions generated by Wound-BLIP. The Grad-CAM visualizations enhance the interpretability of the image-text matching process, offering valuable insights into wound characteristics.

\subsubsection{Exploring an Inter-class Similarity Example}

\begin{figure*}[!th]
  \centering
  \includegraphics[width=1\linewidth]{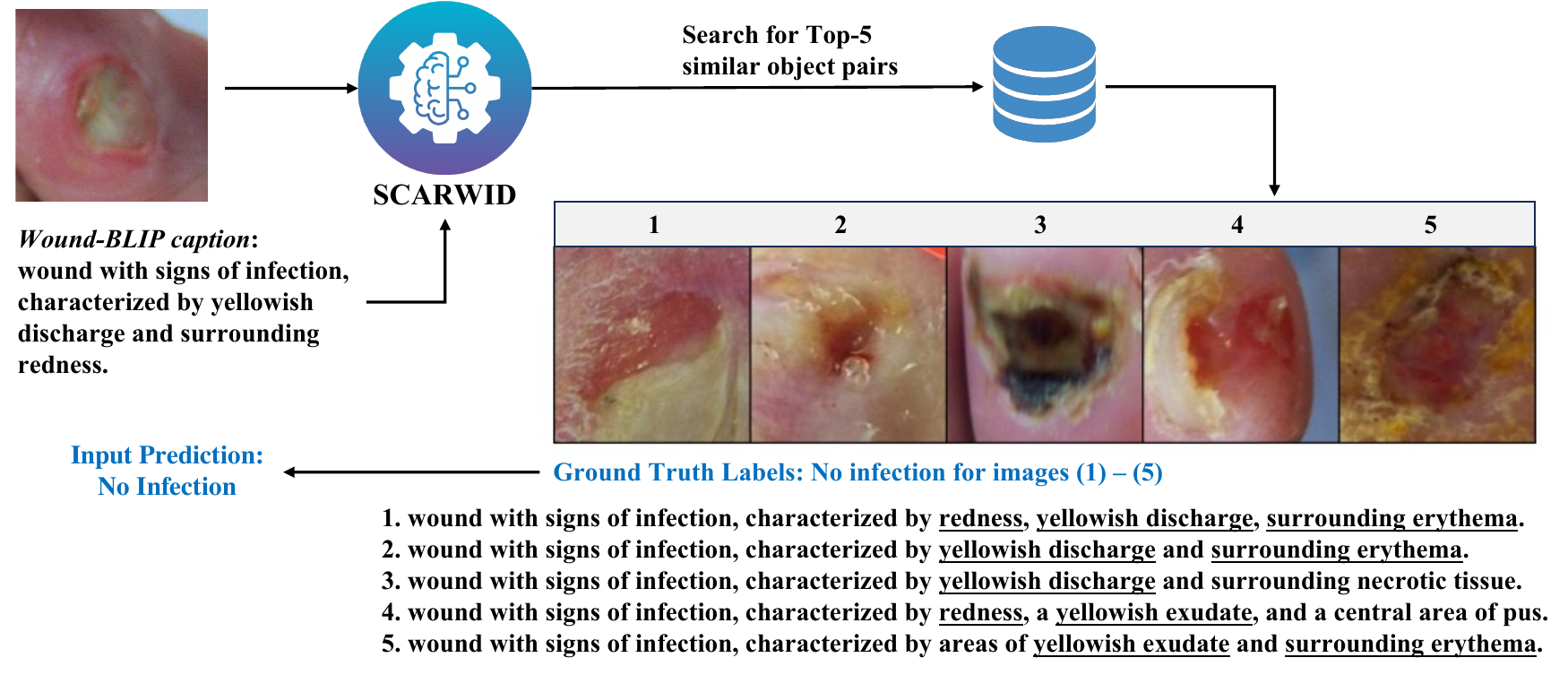}
  \caption{An example of challenging infection detection in a DFU image by SCARWID. Where the similar image-text pairs retrieved from the support collection are represented in numbers (1) to (5).}
  \label{fig:inter-class-pred}
\end{figure*}

As mentioned in Sec.~\ref{sec:Performance}, sometimes, uninfected and infected wounds have very similar visual appearances making it difficult to accurately diagnose wound statuses just from images. Fig.~\ref{fig:inter-class-pred} shows the case where an uninfected wound was described as showing possible signs of infection by our Wound-BLIP captioning model while the input image shows a visual appearance similar to \text{yellowish discharge}. However, text alone is not adequate to make a final decision. Instead, our SCARWID framework tries to find similar pairs of images and texts from the support data collection. Consequently, since the retrieved images were all labeled as uninfected, the input wound is then classified as uninfected even though their corresponding wound captions also present features that possibly appear in infected wounds.

\subsubsection{Exploring Misclassifications} 

 \begin{figure*}[!th]
  \centering
  \includegraphics[width=0.9\linewidth]{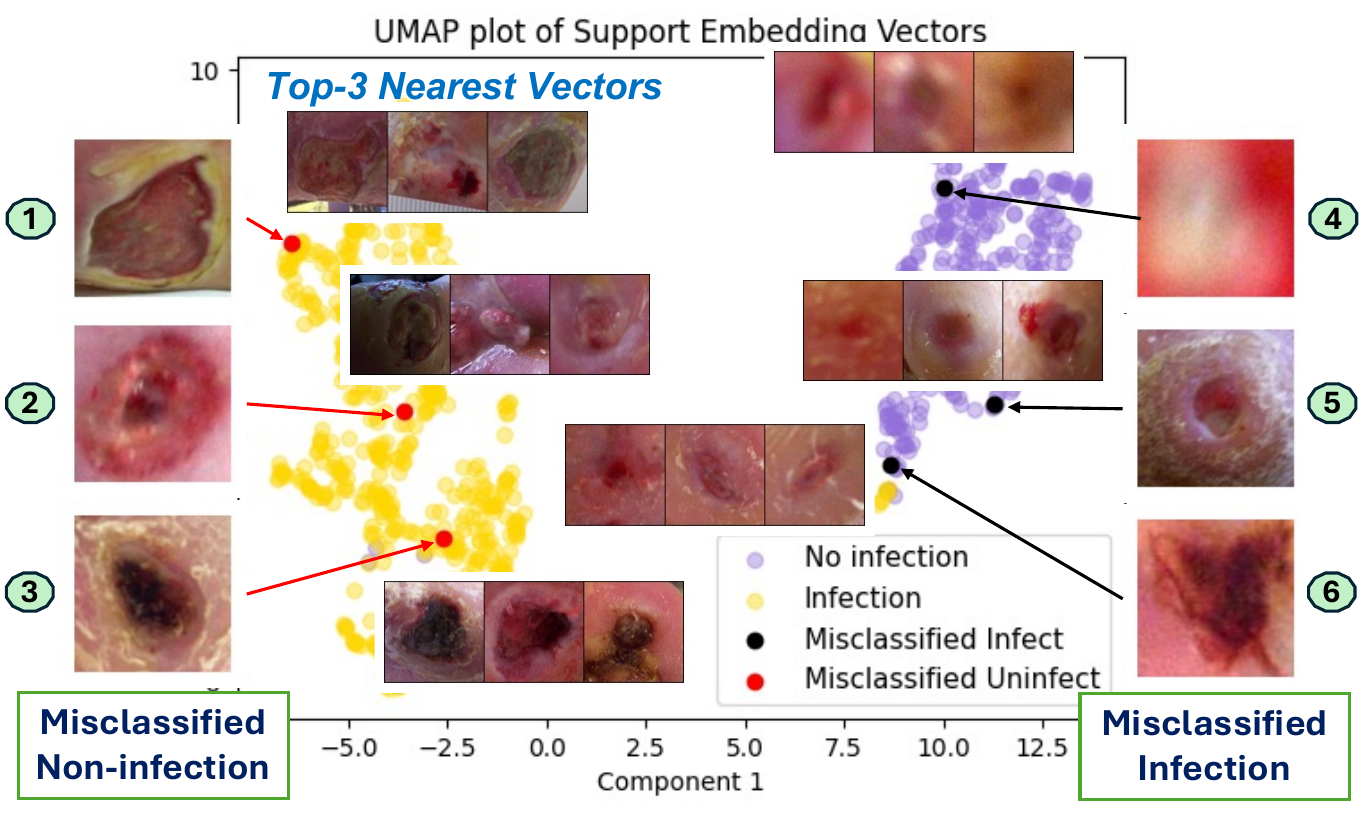}
  \caption{Examples of incorrectly classified DFU images for
infection detection by the SCARWID model. The red points in the UMAP plot indicate misclassified uninfected wounds and the black points indicate misclassified infected wounds. For each incorrectly classified image, the corresponding top-3 similar images are also illustrated.}
  \label{fig:error_analysis}
\end{figure*}

\textbf{The uninfected DFUs in Fig.~\ref{fig:error_analysis} (1-3) that were misclassified as infected wounds exhibit characteristics typically associated with infections}, such as significant reddening or darkening of the tissue. For example, wound 1 shows a yellowish exudate and surrounding erythema, which frequently appear in infected wounds. Wound 2 has redness and the presence of potential pus, leading the model to predict infection incorrectly. Similarly, wound 3 displays an area that appears necrotic, a characteristic also found in infected wounds.

\textbf{The infected wounds in Fig.~\ref{fig:error_analysis} (4-6) that were misclassified as uninfected stem from factors such as small wound size, poor image quality, and ambiguous features}, such as a somewhat dry appearance. Notably, Wound 6, was placed in the uninfected cluster, and is described by the Wound-BLIP model as a close-up of a wound having a dark central area surrounded by reddened skin, indicative of possible infection and inflammation. This highlights a discrepancy where the description appears somewhat accurate, yet the supporting images do not align correctly with the expected class. This discrepancy may result from the presence of fine-grain wound characteristics making it challenging for a VLM to generate accurate response.

\section{Discussion}
\label{sec:Discussion}

\underline{Summary of Findings}:
\textbf{Our proposed SCARWID outperforms other deep learning models in detecting infections in DFU images} by incorporating textual wound descriptions generated by a fine-tuned Vision Language model.  Unlike traditional probability-based models, SCARWID employs a similarity-based approach, leveraging labels from objects retrieved from a labeled support data collection. This methodology enables the model to broaden its search region, effectively handling high intra-class variation in the images.

\textbf{Data augmentation using high-quality synthetic images generated using a latent diffusion model significantly enhances model performance}  on infection classification from DFU images, especially SOTA transformer-based models such as DeiT, SwinV2, and EfficientFormer. This suggests the possibilty of broader applications of similar augmentation techniques in other areas of wound and medical image analysis.

\textbf{Cross-attention mechanism yields more accurate embedding vectors than embeddings on individual image or text modalities, addressing high inter-class similarity.} The Image-Text Fusion module's cross-attention mechanism offers substantial benefits, enhancing the model's capability to interpret and integrate information effectively, resulting in more accurate embedding vectors that address the challenge of high inter-class similarity.

\textbf{Heatmap of Rollout Attention enhances interpretability, demonstrating that SCARWID focuses on the most relevant wound features.} Specifically, it shows that the image encoder in SCARWID focuses on relevant wound features that are described textually, which is particularly useful for Image-Text retrieval process. Thus, it validates SCARWID's interpretative and decision-making processes.

\textbf{Grad-CAM visualization of the alignment of wound characteristics generated by Wound-BLIP and wound features in an image} provides multi-modal wound-related contextual information that can be cross-examined by caregivers and relate the SCARWID model's detection to their medical knowledge.

\underline{Limitations}:
\textbf{Erroneous wound descriptions due to hallucination.} The primary drawback of our framework is the occasional incorrect Wound-BLIP image descriptions, likely due to inaccuracy in the GPT-4o-generated data used for fine-tuning Wound-BLIP. In medical visual question-answering tasks, Multimodal Large Language Models (MLLMs) can sometimes produce textual hallucinations, which refer to misalignment between generated response and actual image content~\cite{hallucination_med, hallucination_sur, yin2023survey}. Consequently, such textual descriptions should not used as standalone medical diagnoses without corroborative image analysis. This underscores the need for future research on mitigating the effects of hallucination of vision-language models in medical contexts. \textbf{Thus, it is necessary to validate the model's descriptive outputs in a study involving medical experts.}

\textbf{Variable quality of wound infection dataset.} The quality of images in the wound infection dataset \cite{partb_DFU} presented a challenge. Some images were blurry and only showed the wound patches. This limitation may have adversely affected the foundation models' ability to precisely locate wounds, subsequently impacting the decision-making process.

\underline{Future Work}: Potential future research directions include applying Retrieval-Augmented Generation (RAG)~\cite{RAG} for MLLMs, enabling them to provide more accurate and evidence-based clinical reasoning corresponding to symptoms through sophisticated search mechanisms~\cite{med-gemini}. Secondly, recent research~\cite{chen2023enhancing} has demonstrated that prompt engineering strategies significantly influence the performance of LLMs in medical tasks. One promising approach is to design prompts that compel GPT-4o to deliver structured responses reflecting three specific capabilities: 1) Image Comprehension, 2) Recall of Medical Knowledge, and 3) Step-by-Step Reasoning before making a final diagnosis~\cite{jin2024hidden}. 

\section{Conclusion}
\label{sec:conclusion}
This paper introduces Synthetic Caption Augmented Retrieval for Wound Infection Detection (SCARWID), a novel multimodal vision-language framework designed to classify infections in diabetic foot ulcers (DFUs) while providing clear, explanatory captions of wound images. These explanations assist novice nurses in recognizing key wound features that are critical for diagnosing infections. SCARWID's combination of a Wound-BLIP model for generating descriptive metadata from DFU images and diffusion-based synthetic image augmentation, significantly enhanced diagnostic capabilities beyond current state-of-the-art methods. SCARWID's innovative use of a labeled support collection's cross-modal embeddings to facilitate a multi-modal retrieval-based classification strategy demonstrated substantial improvements in infection detection accuracy, achieving 81.4\% on a diverse and challenging DFU dataset. This performance not only highlights the efficacy of combining vision and language models in a unified framework but also showcases the potential for this approach to be adapted for other complex medical imaging tasks. Moreover, the improved performance by using a latent diffusion model for image augmentation opens up new avenues to enhance the robustness of AI applications in medical settings. 

\section*{Acknowledgment}
This work is supported by the National Institutes of Health (NIH) through grant 1R01EB031910-01A1 Smartphone-based wound infection screener by combining thermal images and photographs using deep learning methods. The experiments were performed using computational resources provided by the Academic \& Research Computing group at Worcester Polytechnic Institute.

\section*{Declaration of competing interest}
The authors declare that they have no conflicts of interest.
\bibliographystyle{ACM-Reference-Format}
\bibliography{main}

\end{document}